\documentclass[lettersize,journal]{IEEEtran}
\usepackage{amsmath,amsfonts}
\usepackage{algorithmic}
\usepackage{algorithm}
\usepackage{array}
\usepackage[caption=false,font=normalsize,labelfont=sf,textfont=sf]{subfig}
\usepackage{textcomp}
\usepackage{stfloats}
\usepackage{url}
\usepackage{verbatim}
\usepackage{graphicx}
\usepackage{cite}
\usepackage{multirow}

\usepackage{times}
\usepackage{epsfig}
\usepackage{amssymb}
\usepackage{enumitem}
\usepackage{wrapfig}
\usepackage[table]{xcolor}
\usepackage{color}
\usepackage{adjustbox}
\usepackage{booktabs}
\usepackage{xltabular}
\usepackage{authblk}

\title{SVT: Supertoken Video Transformer for Efficient Video Understanding}
\author[1]{Chenbin Pan}
\author[2]{Rui Hou}
\author[2]{Hanchao Yu}
\author[2]{Qifan Wang}
\author[1]{Senem Velipasalar}
\author[2]{Madian Khabsa}
\affil[1]{Syracuse University}
\affil[2]{Meta AI}

\begin{document}

\markboth{Journal of \LaTeX\ Class Files,~Vol., No., 2022}%
{Shell \MakeLowercase{\textit{et al.}}: A Sample Article Using IEEEtran.cls for IEEE Journals}


\maketitle

\begin{abstract}
Whether by processing videos with fixed resolution from start to end or incorporating pooling and down-scaling strategies, existing video transformers process the whole video content throughout the network without specially handling the large portions of redundant information. In this paper, we present a Supertoken Video Transformer (SVT) that incorporates a Semantic Pooling Module (SPM) to aggregate latent representations along the depth of visual transformer based on their semantics, and thus, reduces redundancy inherent in video inputs.~Qualitative results show that our method can effectively reduce redundancy by merging latent representations with similar semantics and thus increase the proportion of salient information for downstream tasks.~Quantitatively, our method improves the performance of both ViT and MViT while requiring significantly less computations on the Kinectics and Something-Something-V2 benchmarks.~More specifically, with our SPM, we improve the accuracy of MAE-pretrained ViT-B and ViT-L by 1.5\% with 33\% less GFLOPs and by 0.2\% with 55\% less FLOPs, respectively, on the Kinectics-400 benchmark, and improve the accuracy of MViTv2-B by 0.2\% and 0.3\% with 22\% less GFLOPs on Kinectics-400 and Something-Something-V2, respectively. 

\end{abstract}


\section{Introduction}
\begin{figure*}[bt!]
  \centering
   \includegraphics[width=1.0\linewidth]{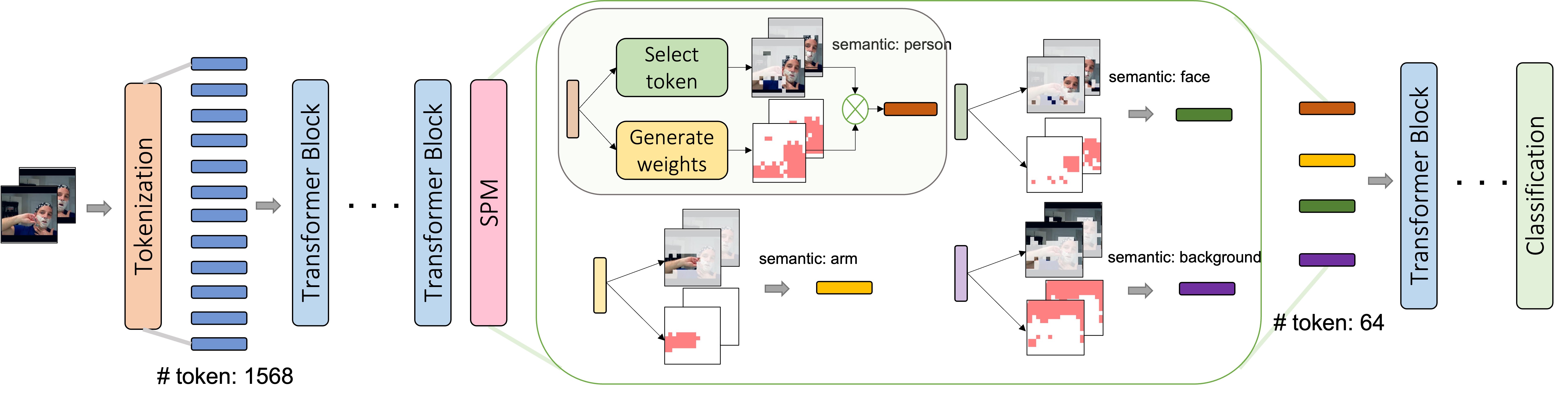}
   \caption{An example of the Supertoken Video Transformer architecture (MAE-ViT-L-SPM18). After the proposed Semantic Pooling Module, the number of tokens is reduced from 1568 to 64.}
   \vspace{-3mm}
   \label{fig:_fig_sempool_arch}
\end{figure*}
Identifying actors or foreground objects in videos is important for the video understanding task due to high redundancy in videos caused by similar backgrounds covering large areas. 
Large computation requirement and longer training times also make the video-related tasks more challenging than images, especially more so for the computationally expensive vision transformers~\cite{vaswani2017attention,devlin2018bert,arnab2021vivit,dosovitskiy2020image,wu2022tinyvit,cho2022cross,gao2022aiatrack,ye2022joint,yan2022towards,zhao2022tracking,dong2022cswin,pu2022edter,wang2022bridged,zhou2021deepvit,ni2022expanding}.
In this work, considering the redundancy and the large semantic overlap between video frames, we propose a Supertoken Video Transformer, referred to as the SVT, which can effectively reduce the redundancy at the semantic level, and thus, significantly decrease the complexity and memory requirement of transformer-based models. At the same time, our module also provides performance improvement over the state-of-the-art (SoTA) models on commonly used video datasets.

Previous video transformers~\cite{li2022mvitv2,arnab2021vivit,liu2021video,neimark2021video,girdhar2021anticipative,zhang2021vidtr,bulat2021space,liang2022vrt,ranasinghe2022self,wang2022bevt,zhang2021token,girdhar2019video,wang2022deformable,kim2022tubeformer,shi2022video,herzig2022object,liu2022learning} process the whole video content throughout the network without specially handling the redundant information. For instance, the Video Vision Transformer (ViViT)~\cite{arnab2021vivit} primarily focuses on modeling global attention on the non-overlapping spatial-temporal video tokens. MViT~\cite{li2022mvitv2,fan2021multiscale} produces multi-scale feature maps by creating a hierarchical architecture with multiple stages from high-resolution to low-resolution. Whether by processing videos with fixed resolution from start to end or incorporating pooling and downscaling strategies, mining features from entire videos, which have highly redundant content, can result in unnecessary computations.
%
In addition, existing works on efficient transformers perform latent representation pooling uniformly according to the fixed space-time shape. In MViT~\cite{li2022mvitv2,fan2021multiscale}, the $qkv$ tensor is generated by a linear layer followed by a $3 \times 3$ kernel group convolution layer applied on the space-time feature maps, and the size of the feature map is controlled by adjusting the stride of the convolution. Swin~\cite{liu2021swin} reduces the size of feature maps by merging adjacent features based on the space-time location. Although this type of uniform downscaling strategies, based on spatio-temporal location, are effective and commonly used to pool features and build hierarchical architectures, they are not conducive to exploring various video semantics with uneven distribution and irregular shapes. 



To address the aforementioned limitations of the existing video transformers, we present a Supertoken Video Transformer (SVT), which uses our proposed Semantic Pooling Module (SPM) to provide adaptability to uneven video information density or video content distribution and avoid unnecessary computation by reducing video redundancy.
In the first line of Fig.~\ref{fig:_fig_tk_dist_cmp1}, we visualize the token distributions of three examples from different classes at the $4^{th}$, $8^{th}$, $12^{th}$, $16^{th}$, $20^{th}$, and $24^{th}$ layers of ViT~\cite{feichtenhofer2022masked}. The red, green and orange colors  represent `swing dancing', `baking cookie' and `golf chipping' classes, respectively. It can be observed that, in the shallow layers ($4^{th}$, $8^{th}$, $12^{th}$), the token distributions are relatively diffused, compared to deeper layers ($16^{th}$, $20^{th}$, $24^{th}$), where the token distribution is more concentrated, which indicates that the token representation learns connections from various low-level features to combined high-level semantics. As the number of tokens representing the similar high-level semantic meanings increases, many computations are unnecessarily performed producing repeated or redundant calculations. Because of such barriers, the three classes cannot be explicitly separated even in the last layer. 
Therefore, based on this observation, we propose to merge features by measuring their distances in multiple semantic spaces as the model goes deeper. Our goal is to effectively save memory space and reduce computations, by removing redundant information, while maintaining semantic diversity. 

\textbf{Contributions.} The main contributions of this work include the following: (i) we present an effective technique, termed Semantic Pooling Module (SPM), to merge latent visual token representations, based on their distances in various learned semantic spaces, into several supertokens;  
(ii) we show that the SPM can be easily applied in both local/global, single/multi-scale video transformers, and achieves great performance-efficiency trade-off;
(iii) we conduct extensive experiments and prove that by adding SPM to both ViT and MViT, comparable or even better performance can be achieved with less computation cost. Specifically,
we improve MAE-ViT-L~\cite{feichtenhofer2022masked} by 0.2\% with 55\% less FLOPs and 0.3\% with 18\% less GFLOPs on the Kinetics-400 benchmark by applying SPM in a hierarchical way and single-pool way, respectively. We improve MViTv2~\cite{li2022mvitv2} by 0.2\% and 0.3\% with 22\% less GFLOPs on Kinetics-400 and Something-Something-V2 datasets, respectively.

\section{Related Work}
\textbf{Video Transformers.} The transformer proposed by Vaswani et al.~\cite{vaswani2017attention} replaces the CNN or RNN layers with self-attention layers, and has been a big success in natural language processing. More recently, Dosovitskiy et al.~\cite{dosovitskiy2020image} proposed a pure Vision Transformer (VIT) for the image classification.~Many other works have focused on building vision transformer models with lower computational cost by using different strategies, such as using semantic visual tokens~\cite{xie2021so}, layer-wise token to token transformation~\cite{yuan2021tokens}, adding distillation losses~\cite{Touvron2021deit} and building a hierarchical structure with the shifted windows~\cite{liu2021swin}.~Video transformers~\cite{Neimark2021VTN, arnab2021vivit,bertasius2021space, liu2021video, patrick2021keeping, herzig2021object, li2022mvitv2, fan2021multiscale} have mirrored the advances in image understanding and SoTA performance on the major video recognition benchmarks~\cite{kay2017kinetics,goyal2017something}. To reduce the computation and memory costs as well as provide locality inductive bias in the self-attention module, Liu et al.~\cite{liu2021video} strictly followed the hierarchy of the original Swin Transformer~\cite{liu2021swin} for the image domain, and extended the scope of local attention computation from only the spatial domain to the spatiotemporal domain.

\textbf{Computation-saving Techniques.}
Sevaral works have been presented for saving computations in transformers \cite{meng2022adavit,yin2022vit,rao2021dynamicvit,kong2021spvit}.
AdaViT~\cite{meng2022adavit} is proposed to adaptively prune tokens throughout the transformer. A-ViT~\cite{yin2022vit} is proposed to adaptively adjust the amount of token computations based on input complexity. Michael et al.~\cite{ryoo2021tokenlearner} propose the tokenlearner for visual representation learning. 
Although it achieves the efficiency goal, pooling tokens globally with the learned weights from the tokenlearner module, the model cannot avoid redundancy and background noise when the core object only occupies a very small area and the obstruction signal is stronger in the video, like the examples in Fig.~\ref{fig:_fig_sempool_visx2}.
In~\cite{bolya2022token}, TokenMerger (ToMe) is proposed to merge similar tokens based on the bipartite matching algorithm. 
In the merging process, the number of tokens in each merging group is added in the softmax of the proportional attention, which cannot handle the situation when there is a larger number of background tokens than the foreground tokens, which more frequently happens in video datasets.
Although they provide efficiency benefits, both tokenlearner and ToMe cannot surpass the ViT baseline, and their usage is limited to the ViT-based models.

\begin{figure}[t]
  \centering
   \includegraphics[width=1.0\linewidth]{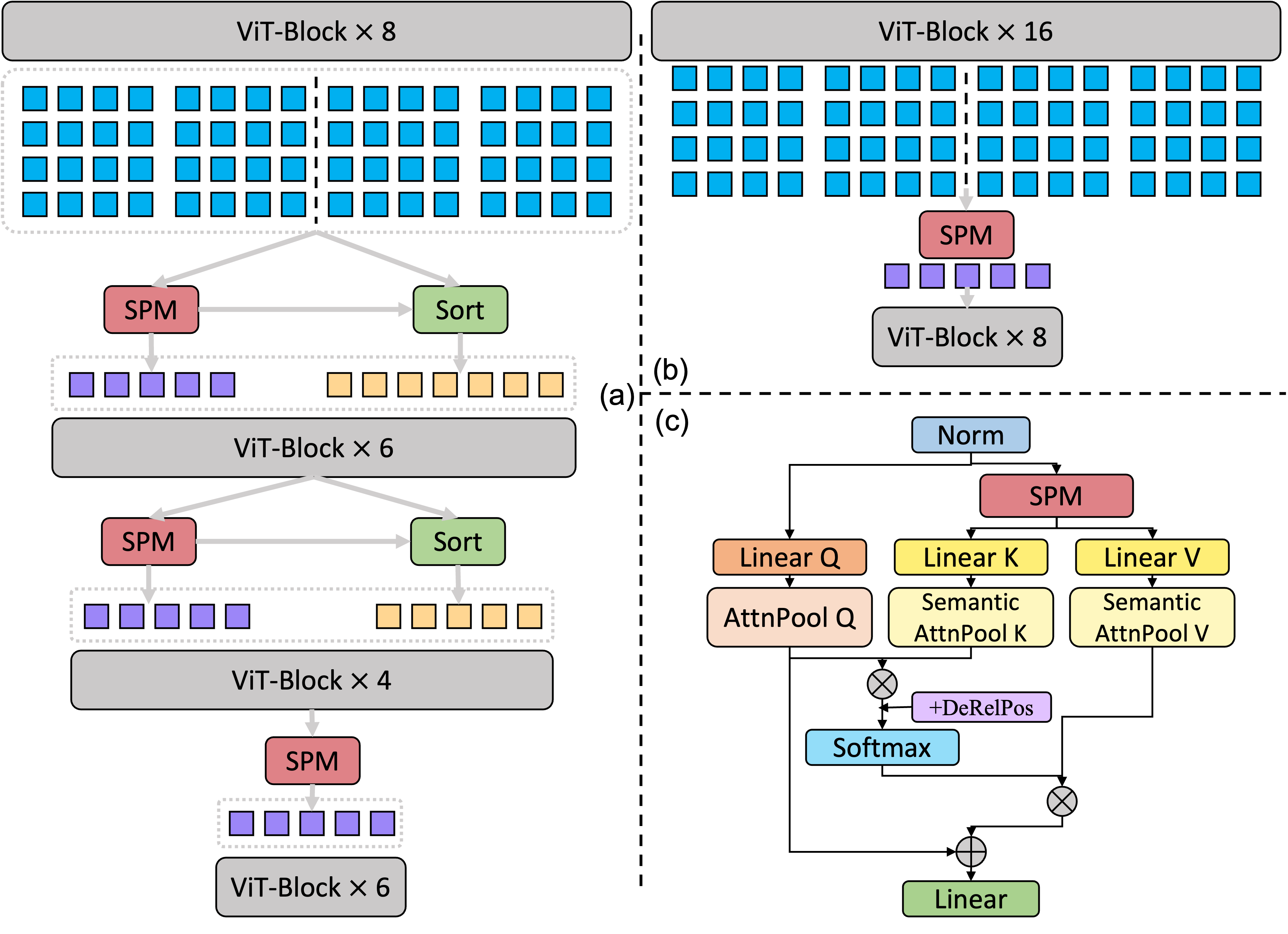}
   \vspace{-0.5cm}
   \caption{(a) The architecture for ViT-L-SPM8/14/18. We insert the SPM in the 8th, 14th, 18th transformer layers to reduce the total number of tokens from 1568 to 1024, 512, and 128, respectively, with 128 semantic tokens after each of pooling; (b) the architecture for ViT-L-SPM/16. We insert the SPM after the 16th layer to reduce the number of tokens from 1568 to 128; (c) multi-scale attention module with SPM in ours-MViTv2.}
   \vspace{-0.6cm}
   \label{fig:_fig_sempool_attn_blk}
\end{figure}

\section{Proposed Model}
\vspace{-0.2cm}
In this section, we present the details of our SVT. We first introduce the preliminaries and main baselines in Sec.~\ref{ssec:Preliminaries}.
Then, we present our SPM and the single-scale/multi-scale building architectures incorporating the SPM in Sec.~\ref{ssec:SPM} and Sec.~\ref{sec:SPM_Model_Arch}, respectively. 
\vspace{-0.2cm}

\subsection{Preliminaries} \label{ssec:Preliminaries}
\vspace{-0.2cm}
We use two latest SoTA models on video task, namely MAE-ViT~\cite{feichtenhofer2022masked} and MViTv2~\cite{li2022mvitv2}, as our single-scale and multi-scale transformer baselines, respectively. In ViT, a 3D video input is segmented and flattened as a 1D sequence tensor of length $L$ and channel $C$, $X \in \mathbb{R} ^ {L \times C} $. The model is constructed of several transformer blocks to perform self-attention and MLP on single-scale feature maps in a global manner without changing its size. 
In the self-attention module, LayerNorm is applied to stabilize the hidden state dynamics of the input tensor. Three linear layers are applied to generate a query,key,value tensor. Scaled matrix multiplication is performed between query and key to produce the attention map, which is normalized by the following SoftMax layer. Another linear layer is applied on the product of attention map and value tensor for output projection. Skip connection is applied to make the information stay local in the transformer layer stack. The MLP module is composed of an MLP layer, a LayerNorm layer, and skip connection. The whole process in a ViT transformer block can be formulated as in Eqs.~(\ref{eq:attn}) and (\ref{eq:mhsa}):
\begin{equation}
\begin{aligned}
  x_1 &= x + MHSA(LayerNorm(x))\\
  x_2 &= x_1 + MLP(LayerNorm(x_1)),
  \label{eq:attn}
\end{aligned}
\end{equation}
\begin{equation}
\resizebox{1\hsize}{!}{$MHSA(x) = FC_{o}(Softmax(\alpha \cdot  FC_{q}(x) \otimes  FC_{k}(x)) \otimes FC_{v}(x))$}.
\label{eq:mhsa}
\end{equation}

MViTv2 is the latest SoTA video transformer with hierarchical architecture for modeling both low- and high-level visual features. The main differences between MViTv2 and ViT are as follows: (i) In the attention mechanism of MViTv2, instead of using three separate single linear layers, three sets of pooling attention (a linear layer followed by a group convolution layer) are used to generate query-key-value tensors and reduce resolution, by enlarging the stride in group convolution layer at the first transformer block of each stage; (ii) the pooling attention allows computing attention on sparse key and value sequences, which helps reduce the computational cost; (iii) in contrast to ViT, where the channel dimension and sequence length remain fixed, MViTv2 progressively increases the channel dimension and reduces spatio-temporal resolution via pooling attention mechanism; (iv) MViTv2 incorporates decomposed relative position embedding and residual pooling connection with the (pooled) query tensor into the self-attention module to enhance the information flow.
The procedure in a MViTv2 transformer block can be expressed as: 
\vspace{-1mm}
\begin{equation}
\begin{aligned}
  q = GConv_{q}(&FC_{q}(x)), \quad k = GConv_{k}(FC_{k}(x)), \\
 &v = GConv_{v}(FC_{v}(x))
\end{aligned}
  \label{eq:convattnpool_qkv}
\end{equation}
\begin{equation}
\begin{aligned}
  attn &= Softmax(\alpha \cdot q \otimes k^{T} + relpos_{k})\\
  Conv&AttnPool(x) = FC_{o}(q + attn \otimes v)\\
  x_1 = & res(x) + ConvAttnPool(LayerNorm(x)) \\
  x_2 = & x_1 + MLP(LayerNorm(x_1))
\end{aligned}
  \label{eq:convattnpool_attn}
\end{equation}



\subsection{Semantic Pooling} \label{ssec:SPM}
Suppose we have $N$ input tokens $\vec{x_j} \in \mathbb{R}^{C}$, where $N$ is the combination of the $T,H,W$ spatio-temporal axes of the video patches. Instead of investigating the pair-wise matching score among the input tokens, we initialize $M$ trainable embeddings, $\vec{e_i} \in \mathbb{R}^{C}$, as explicit semantic prototypes to be used for rating video tokens under different schemes.
Then, the semantic matching degree among the input tokens is estimated by their dot products to these prototypes. 
The produced score map $S \in \mathbb{R}^{M \times N}$ is then sent into an elitism function $\mathfrak{F}(\cdot)$ to select and cluster analogous individuals.
\begin{equation}
\begin{aligned}
  s_{i,j} &= \vec{x_j} \cdot \vec{e_i}\\
   \hat{s}_{i,j} = \mathfrak{F}(s_{i,j}, \psi ,\theta ) &= \begin{cases}
      s_{i,j}, & \text{ if } \psi(s_{i,j}) > \theta \\
      -inf, & \text{ if } \psi(s_{i,j}) \le \theta
    \end{cases} \\
    Z_{i,k} = &Softmax(\hat{S}_{i,k}) \otimes  X_{i,k} \in \mathbb{R}^{1 \times C}
 \end{aligned}
  \label{eq:elitism}
\end{equation}



In the elitism procedure, as expressed by Eq.~(\ref{eq:elitism}),  we apply a non-linear function $\psi(\cdot)$ to compress the affinity score $s_{i,j}$ into the range of $0-1$, where $i$ and $j$ indicate the identity of the semantic space and input token, respectively. We set a fixed threshold $\theta$ as a filter to limit the interactions between input tokens and semantic spaces, i.e., if the compressed score $\psi(s_{i,j})$ is higher than the threshold $\theta$, then its original value $s_{i,j}$ and the corresponding token $x_j$ are preserved in the $i^{th}$ semantic group, while other tokens and values with compressed scores lower than the threshold are muted. In our implementation, we use $sigmoid$ as the compression function.
Therefore, each semantic group can consist of different numbers of active tokens. Many videos can have explicit bias to some semantic groups, and have less content belonging to other semantics. This may lead to semantic groups having no active tokens, which, in turn, causes the gradient vanishing problem during training. Considering this, we set all the tokens with their corresponding semantic scores as active states for such groups. 

In order to enable hierarchical structure, we further split the tokens together with their semantic scores under each prototype into rough local groups. With window size of $T_w \times H_w \times W_w$, the tokens $X \in \mathbb{R}^{T \times H \times W \times C}$ and the semantic scores $S \in \mathbb{R}^{M \times T \times H \times W}$ are segmented into the shapes of $N_{win} \times T_w \times H_w \times W_w \times C$ and $M \times N_{win} \times T_w \times H_w \times W_w$, respectively,
where $N_{win} = N_t \times N_h \times N_w = \frac{T}{T_w} \times \frac{H}{H_w} \times \frac{W}{W_w}$ indicates the number of windows. With various shapes of active semantic areas, the semantic scores are normalized by the $Softmax$ function along the window axis. We then perform the weighted sum over the normalized score and the tokens window-wise. Hence, under $M$-many semantics, the number of tokens with dimension $C$ is reduced to $N_{win}$ in each group, and the total number of output tokens is $M \times N_{win}$. We conduct ablation studies to investigate the trade-off between the number of semantics $M$ and the number of windows $N_{win}$ in Sec.~\ref{ssec:ablations}.

\subsection{Model Architecture} \label{sec:SPM_Model_Arch}
We design different strategies for incorporating our proposed semantic pooling module (SPM) with single-scale transformer ViT and multi-scale transformer MViTv2.

\noindent \textbf{Semantic pooling with ViT.} 
For the combination with ViT, we directly insert the SPM between transformer layers. We propose two different ways of integration as shown in Fig.~\ref{fig:_fig_sempool_attn_blk}(a)(b). 
In Fig.~\ref{fig:_fig_sempool_attn_blk}(a), we insert SPM in an hierarchical way to reduce the number of tokens progressively.
As indicated in Sec.~\ref{ssec:SPM}, we extract a total of $M \times N_{win}$-many semantic tokens after SPM. To combine SPM with ViT in an hierarchical way, in each instance of SPM (except the last one), we keep the top $N_{k}$ original tokens based on their average semantic score. This is done to avoid losing details from the shallow layers of ViT. Hence, after SPM, the total number of tokens is reduced to $N_{r} = M \times N_{win} + N_{k}$.
In our ablation studies in Sec.\ref{ssec:ablations}, we compare different ways of reducing the number of tokens to $N_r$, more specifically, by using  (i) only the semantic tokens, (ii) only the original tokens, and (iii) combination of semantic and original tokens. We show that the combination approach provides the best performance. In Fig.~\ref{fig:_fig_sempool_attn_blk}(b), the SPM is inserted as a single-pool layer to reduce the number of token to $M \times N_{win}$ without keeping the original tokens. In this paper, we construct and investigate several variants for ViT-SPM, as provided in Tab.~\ref{tab:config}, while the effective configurations are not necessarily limited to the examples shown in the table.

\begin{table}
  \begin{adjustbox}{width=0.8\width,center}
  \begin{tabular}{c|ccc}
      \toprule
      Model & L & $T_w\times H_w \times W_w$ & $N_{win} \times M+N_k=N_r$ \\
      \midrule
      MAE-ViT-B-SPM6 & 6 & 2$\times$14$\times$14 & 4$\times$32+0=128 \\
      MAE-ViT-B-SPM8 & 8 & 2$\times$14$\times$14 & 4$\times$32+0=128 \\
      \midrule
      MAE-ViT-L-SPM12 & 12 & 2$\times$14$\times$14 & 4$\times$32+0=128 \\
      MAE-ViT-L-SPM16 & 16 & 2$\times$14$\times$14 & 4$\times$32+0=128 \\
      MAE-ViT-L-SPM18 & 16 & global & 1$\times$64+0=64 \\
      \midrule
      \multirow{3}{*}{MAE-ViT-L-SPM8/12/16} & 8 & 2$\times$14$\times$14 & 4$\times$32+896=1024 \\
                                          & 12 & global & 1$\times$128+384=512 \\
                                          & 16 & global & 1$\times$128+0=128 \\
      \midrule
      \multirow{3}{*}{MAE-ViT-L-SPM8/14/18} & 8 & 2$\times$14$\times$14 & 4$\times$32+896=1024 \\
                                          & 14 & global & 1$\times$128+384=512 \\
                                          & 18 & global & 1$\times$128+0=128 \\
      \bottomrule
  \end{tabular}
  \end{adjustbox}
  \caption{\small{Model variants of MAE-ViT-SPM. L indicates the layer where SPM is inserted, $T_w\times H_w \times W_w$ indicates the window size in SPM, and $N_r$ is the total number of tokens after SPM.}} 
  \vspace{-0.6cm}
  \label{tab:config}
\end{table}

\noindent \textbf{Semantic pooling with MViTv2.} We design a semantic attention module for the multi-scale transformer MViTv2, as shown in Fig.~\ref{fig:_fig_sempool_attn_blk}(c). We replace the original multi-scale attention module with the semantic one for every 4 blocks in our experiments.
For input $X$, with shape $T \times H \times W \times C$, we apply SPM with window size of $T_w \times H_w \times W_w$ and $M$-many
semantic groups to generate semantic tokens $X_{sem} \in \mathbb{R}^{N_{win}^{T} \times N_{win}^{H} \times N_{win}^{W} \times M \times C}$. We apply two separate linear layers on $X_{sem}$ to generate intermediate tensors for key and value. Then, each of them is sent into a $3 \times 3$ group convolution with the number of group equal to $M \times C$ and stride equal to 1 to produce the final key/value tensor 
as shown in Eq.~(\ref{eq:q_k_v_tensor}). Hence, the key-value pairs are focusing on the semantic content, while the query is based on the spatial-temporal map. 
Then, each query will attend to the semantic spaces by computing the pair-wise attention with $K_{sem}$. The whole process of our semantic attention module can be expressed as follows: 
\begin{equation}
\begin{aligned}
X_{sem} &= SPM(X) \in \mathbb{R}^{N_{win}^{T} \times N_{win}^{H} \times N_{win}^{W} \times M \times C},\\
Q &= GConv_{q}(FC_{q}(X)) \in \mathbb{R}^{T \times H \times W \times C}, \\
K_{sem} &= GConv_{k}(FC_{k}(X_{sem})) \in \mathbb{R}^{N_{win} \times M \times C}, \\
V_{sem} &= GConv_{v}(FC_{v}(X_{sem})) \in \mathbb{R}^{N_{win} \times M \times C}, \\
A_{sem} &= Softmax(\alpha \cdot Q \otimes K_{sem}^{T} + relpos_{k}),\\
X_{out} &= FC_{o}(Q + A_{sem} \otimes V_{sem}), 
  \label{eq:q_k_v_tensor}
\end{aligned}
\end{equation}
where $A_{sem}$ is the attention map between $Q$ and $K_{sem}$ and $X_{out}$ is the final output tensor.
Therefore, in the original multi-scale attention module in MViTv2, the stride in K/V attention pool can be larger than the kernel size causing middle patches to be ignored.
In contrast, in our semantic attention module, all patches will contribute to K/V tensor via SPM, avoiding the risk of losing information. Also, with SPM, the number of K/V tokens can be reduced to a smaller number, making the module more efficient.
\vspace{-0.3cm}


%


\section{Experiments}\label{sec:4_exp}
\vspace{-0.2cm}
\begin{figure*}[bt!]
  \centering
   \includegraphics[width=1.0\linewidth]{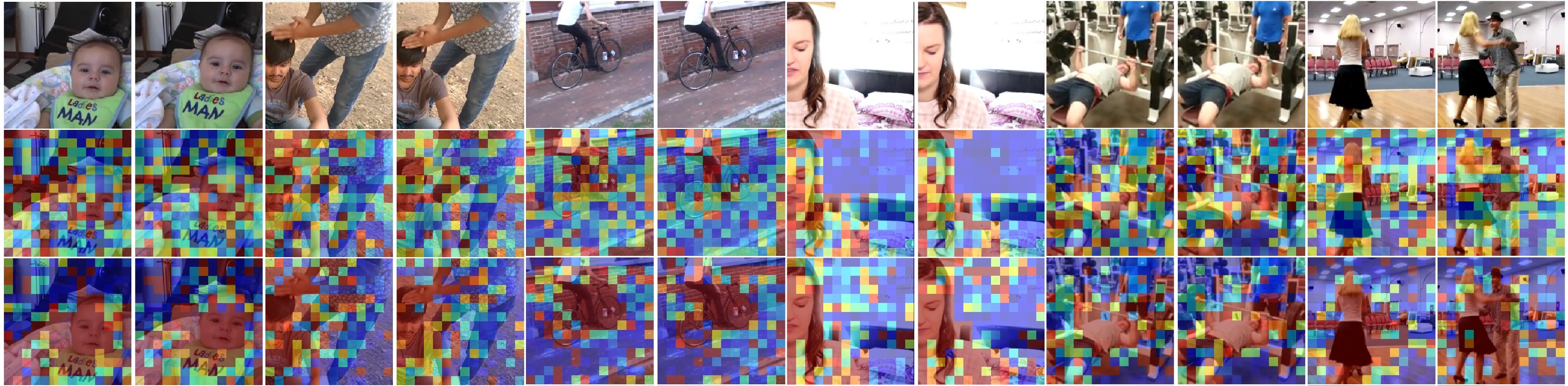}
   \caption{Visualization of attention comparison between MAE-ViT-L (baseline) and MAE-ViT-L-SPM16 (ours). Classes from left to right are: `sneezing', `massaging person's head', `riding a bike', `tasting food', `benching pressing', and `swing dancing', respectively. We show the raw video frames, the averaged attention score for each patch/token in the $16^{th}$ layer of the baseline, and the averaged semantic score for each patch/token in the $16^{th}$ layer of our method (during the SPM) in the first, second, and third rows of each example, respectively. The redder the token is, the higher attention is placed on it.}
   \label{fig:iccv_attn_vis}
   \vspace{-0.5cm}
\end{figure*}

We conduct extensive experiments on Kinetics and SSv2 datasets for video recognition to validate our model. We divide the experiments into two groups, wherein single-scale ViT and multi-scale MViTv2 are applied as baselines, respectively. In the first group, both ViT and our models adopt the same pre-trained weights from the MAE-ViT, while in the second group, both MViTv2 and our models are trained from scratch. We show and discuss the results of these two sets of experiments separately, and further provide the qualitative examples for comparison.



\vspace{-0.2cm}
\subsection{Video Recognition}
\vspace{-0.2cm}
\noindent \textbf{Kinetics-Settings.}~Kinetics-400 (K400) is a large-scale video dataset including 400 human action classes, with at least 400 video clips for each action. We train the models following the recipes provided in \cite{feichtenhofer2022masked, li2022mvitv2}. For ViT-based models, we adopt the same pre-trained weights from the MAE-ViT \cite{feichtenhofer2022masked}, while for MViTv2 based models, we train from scratch.

\noindent \textbf{Kinetics-Results.}~Table~\ref{tab:k400} shows our results on K400. 
We show the results of MAE-ViT based models in the first three groups, and typical multi-scale transformers in the last two groups.
For all the base, large, and huge versions of MAE-ViT, SPM not only helps improving the performance but also reduces the computation and memory requirements. 
The MAE-ViT-B-SPM6/SPM8 provides 0.6\%/1.5\% performance improvement, and 50\%/33\% decrease on the number of GFLOPs. MAE-ViT-L-SPM18/SPM8/14/18 provides 0.3\%/0.2\% performance improvement, and 55\%/18\% decrease on the number of GFLOPs.
In addition, compared to other computation saving techniques~ MAE-ViT-Tokenlearner~\cite{ryoo2021tokenlearner} and ToMe~\cite{bolya2022token}, ours-B-SPM6/SPM8 outperforms the Tokenlearner-B by 1.7\%/2.6\% accuracy increase with smaller computations; ours-L-SPM18/SPM8/14/18 surpasses the Tokenlearner-L by 0.6\%/0.5\% accuracy increase with 56\%/75\% less number of GFLOPs; and ours-L-SPM8/14/18 surpasses the ToMe-L by 0.6\% accuracy with less computations.
While other computation saving techniques can only be applied in ViT-based model, our SPM can also be applied in multi-scale transformers. As shown in Tab.\ref{tab:k400}, for MViTv2-based models, ours-S-SPM can achieve comparable performance with 5 less GFLOPS than MViTv2-S, and ours-B-SPM outperforms the MViTv2-B by 0.2\% accuracy with 22\% less GFLOPs.
Both results based on MAE-ViT and MViTv2 prove the effectiveness and efficiency of our proposed SPM.

\begin{table}[t]
\centering
\resizebox{1.0\linewidth}{!}{
\begin{tabular}{lcccc}
  \toprule
  model & top-1 & top-5 & GFLOPs & \#Params/M \\
  \midrule
  MAE-ViT-B$\dagger$ & 79.3 & 93.2 & 180$\times$3$\times$7 & 87 \\
  Tokenlearner-B$\dagger$ & 78.2 & 93.2 & 120$\times$3$\times$7 & 87 \\
  \rowcolor{blue!10}ours-B-SPM6 & \textbf{79.9}(\textcolor{red}{+0.6}) & \textbf{94.5} & 91$\times$3$\times$7(\textcolor{red}{$\downarrow$50\%}) & 87 \\
  \rowcolor{blue!10}ours-B-SPM8 & \textbf{80.8}(\textcolor{red}{+1.5}) & \textbf{94.8} & 120$\times$3$\times$7(\textcolor{red}{$\downarrow$33\%}) & 87 \\
  \midrule

  MAE-ViT-L\cite{feichtenhofer2022masked} & 84.8 & 96.2 & 598$\times$3$\times$7 & 304 \\
  TokenLearner-L\cite{bolya2022token} & 84.5 & - & 1105$\times$4$\times$3 & 383 \\
  ToMe-MAE-ViT-L\cite{bolya2022token} & 83.2 & - & 184$\times$1$\times$10 & 304 \\
  ToMe-MAE-ViT-L\cite{bolya2022token} & 84.4 & - & 281$\times$1$\times$10 & 304 \\
  \rowcolor{gray!20}ours-L-SPM12 & 84.6 & 96.2 & 302$\times$3$\times$7 & 304 \\
  \rowcolor{blue!10}ours-L-SPM16 & \textbf{85.0}(\textcolor{red}{+0.2}) & \textbf{96.6} & 473$\times$3$\times$7(\textcolor{red}{$\downarrow$21\%}) & 304 \\
  \rowcolor{blue!10}ours-L-SPM18 & \textbf{85.1}(\textcolor{red}{+0.3}) & \textbf{96.5} & 490$\times$3$\times$7(\textcolor{red}{$\downarrow$18\%}) & 304 \\
  \rowcolor{gray!20}ours-L-SPM8/12/16 & \textbf{84.8} & \textbf{96.4} & 254$\times$3$\times$7(\textcolor{red}{$\downarrow$58\%}) & 304 \\
  \rowcolor{blue!10}ours-L-SPM8/14/18 & \textbf{85.0}(\textcolor{red}{+0.2}) & \textbf{96.5} & 275$\times$3$\times$7(\textcolor{red}{$\downarrow$55\%}) & 304 \\
  \midrule
  MAE-ViT-H\cite{feichtenhofer2022masked} & 85.1 & 96.6 & 1193$\times$3$\times$7 & 632 \\
  \rowcolor{gray!20}ours-H-SPM22 & \textbf{85.1} & \textbf{96.7} & 955$\times$3$\times$7(\textcolor{red}{$\downarrow$20\%}) & 632 \\
  \midrule
  \midrule
  MViTv1, 16x4\cite{fan2021multiscale}& 78.4 & 93.5 & 70$\times$1$\times$5 & 37 \\
  Swin-S, (IN1K)\cite{liu2021video} & 80.6 & 94.5 & 166$\times$4$\times$3 & 50 \\
  MViTv2-S, 16x4\cite{li2022mvitv2} & 81.0 & 94.6 & 64$\times$1$\times$5 & 34 \\
  \rowcolor{gray!20}ours-S-SPM, 16x4 & 80.9 & \textbf{94.6} & 53$\times$1$\times$5(\textcolor{red}{$\downarrow$17\%}) & 34 \\
  \midrule
  MViTv1, 32x3\cite{fan2021multiscale}& 80.2 & 94.4 & 170$\times$1$\times$5 & 37 \\
  Swin-B, (IN1K)\cite{liu2021video} & 80.6 & 94.6 & 282$\times$4$\times$3 & 88 \\
  Swin-B, (IN21K)\cite{liu2021video} & 82.7 & 95.5 & 282$\times$4$\times$3 & 88 \\
  MViTv2-B, 32x3\cite{li2022mvitv2}& 82.9 & 95.7 & 225$\times$1$\times$5 & 51 \\
  \rowcolor{blue!10}ours-B-SPM, 32x3 & \textbf{83.1}(\textcolor{red}{+0.2}) & 95.6 & 176$\times$1$\times$5(\textcolor{red}{$\downarrow$22\%}) & 51 \\
  \bottomrule
\end{tabular}
}
\caption{\small{Video recognition results on the K400 dataset. For the MAE-ViT-B labeled with $\dagger$, we report the result from our own reproduction, which is lower than the number stated in the original paper. We also report our own reproduction for Tokenlearner-B labeled with $\dagger$ using the MAE-pretrained weights, since there is no such result in the original paper \cite{ryoo2021tokenlearner}. We mark the results from our proposed model in gray lines and the best performances in blue lines. The first three groups of the table present the comparisons among the single-scale transformers based on MAE-ViT, while the last two groups show the comparisons among the multi-scale transformers.}}
\label{tab:k400}
\vspace{-0.6cm}
\end{table}

\noindent \textbf{SSV2-Settings.}~Something-Something-v2 (SSV2) is a collection of 220,847 labeled video clips covering 174 classes of humans performing actions with everyday objects. We train the models following the recipes provided in \cite{feichtenhofer2022masked, li2022mvitv2}. For MViTv2-based models, we utilize the pre-trained weights from K400 task to initialize the corresponding models, while for ViT-based models, we utilize the same pre-trained weights as in K400 task.

\noindent \textbf{SSV2-Results.} We provide the video classification results for both MAE-ViT- and MViTv2-based models on SSV2 in Tab.~\ref{tab:ssv2}. It can be seen that, with fewer computations, each of our models achieve competitive  if not better performance compared to the corresponding baseline model. Our SPM also shows superior feature exploring ability than other efficiency-oriented techniques, which invariably degrade model performance.
This preeminence can be attributed to the following: with much redundancy in videos, while the baselines uniformly distribute tokens among the whole video content, our model can adjust the token distribution by increasing the portion of tokens representing the core object(s) and decreasing the portion of tokens representing the background and irrelevant objects (token distribution visualization details are provided in Sec.~\ref{ssec:qualitative} and Fig.~\ref{fig:_fig_tk_dist_cmp1}). Proposed SPM, while improving the foreground ratio in the token pools, does not entirely wipe out the background information. This implies that it does not lose information while removing redundancy. Furthermore, SPM serves as a flexible aggregator that groups tokens in a data-driven way without fixing the number of tokens in each pooling group. Benefiting from the internal logic of pooling and adjusting the token distribution based on semantic meaning, our model can boost the performance and decrease computation and memory requirements at the same time.

\noindent \textbf{Accuracy-efficiency tradeoff.} We compared the accuracy-efficiency tradeoff of our SVT with other most-recently published SOTAs on K400 dataset in the Fig.\ref{fig:_fig_tk_dist_cmp1}.
We analyze the reasons for that our SVT achieves better accuracy-efficiency trade-off than other computation saving techniques are: (i) When reducing tokens, the Tokenlearner\cite{ryoo2021tokenlearner} captures global information in each token, while ours captures semantic information, including both global and local instances, which can better explore the interrelationship of the core objects in the video.
Tokenlearner focuses on re-weighting the whole scene to generate tokens, while ours focuses on adjusting the ratio of foreground to background in the token pool, and gathering similar semantics to generate supertokens, which is a better way to explore the intrinsic features of the video as proved in our experiment results; (ii)
In the bipartite matching process of ToMe \cite{bolya2022token}, the number of foreground and background tokens is decreased simultaneously with equal proportions, while in our SPM, we improved the ratio of foreground tokens, as shown in Fig.\ref{fig:iccv_attn_vis} and our supplement examples.

\begin{table}[t]
\centering
\resizebox{1.0\linewidth}{!}{
\begin{tabular}{lcccc}
  \toprule
  model & top-1 & top-5 & GFLOPs & \#Params/M \\
  \midrule
  MAE-ViT-L\cite{feichtenhofer2022masked} & 72.1 & 93.9 & 598$\times$3$\times$1 & 304 \\
  \rowcolor{gray!20}ours-L-SPM16 & 71.7 & 93.8 & 473$\times$3$\times$1(\textcolor{red}{$\downarrow$21\%}) & 304 \\
  \rowcolor{gray!20}ours-L-SPM18 & 71.9 & \textbf{94.0} & 490$\times$3$\times$1(\textcolor{red}{$\downarrow$18\%}) & 304 \\
  \midrule
  MAE-ViT-H\cite{feichtenhofer2022masked} & 74.1 & 94.5 & 1193$\times$3$\times$1 & 632 \\
  \rowcolor{gray!20}ours-H-SPM22 & 73.6 & \textbf{94.6} & 955$\times$3$\times$1(\textcolor{red}{$\downarrow$20\%}) & 632 \\
  \midrule
  \midrule
  MViTv1, 16x4\cite{fan2021multiscale} & 64.7 & 89.2 & 71$\times$1$\times$5 & 37 \\
  MViTv2-S, 16x4\cite{li2022mvitv2} & 68.2 & 91.4 & 64$\times$1$\times$5 & 34 \\
  \rowcolor{gray!20}ours-S-SPM, 16x4 & 68.0 & 91.0 & 53$\times$1$\times$5(\textcolor{red}{$\downarrow$17\%}) & 34 \\
  \midrule
  MViTv1, 64x3\cite{fan2021multiscale} & 67.7 & 90.9 & 454$\times$1$\times$5 & 37 \\
  MViTv2-B, 32x3\cite{li2022mvitv2} & 70.5 & 92.7 & 225$\times$1$\times$5 & 51 \\
  \rowcolor{blue!10}ours-B-SPM, 32x3 & \textbf{70.8}(\textcolor{red}{+0.3}) & \textbf{93.0} & 176$\times$1$\times$5(\textcolor{red}{$\downarrow$22\%}) & 51 \\
  \bottomrule
\end{tabular}
}
\vspace{-0.3cm}
\caption{Video Recognition Results on SSV2 Dataset.}
\label{tab:ssv2}
\vspace{-0.6cm}
\end{table}

\subsection{Qualitative Results} \label{ssec:qualitative}
In this section, we provide some visualization examples to better illustrate benefits of our proposed SPM. More visualizations are provided in the supplementary material.

\begin{figure*}[bt!]
    \centering
    \includegraphics[width=1.0\linewidth]{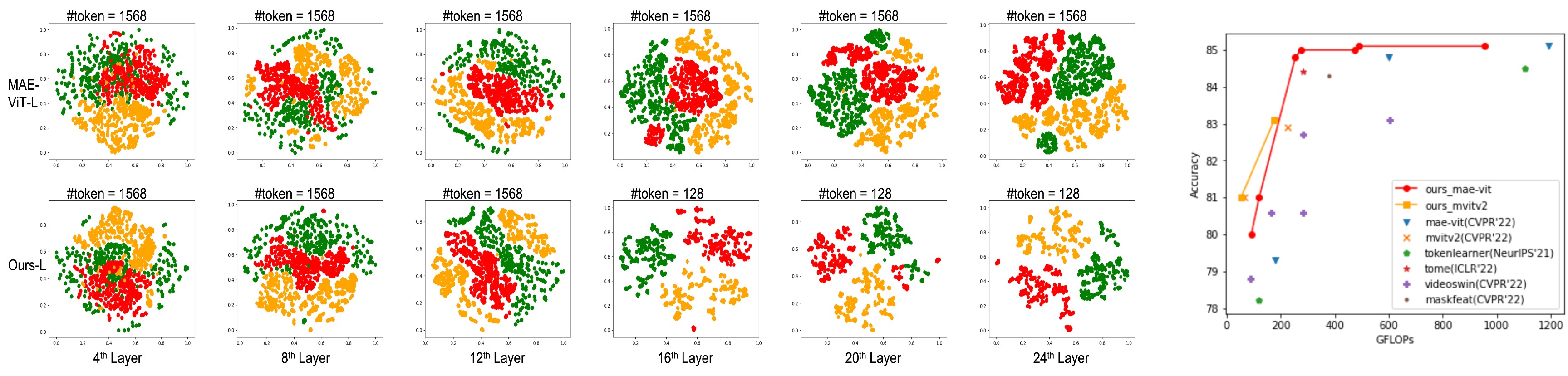}
    \vspace{-0.8cm}
    \caption{Left: Visualization of latent token representations after different layers along the depth of MAE-ViT-L and MAE-ViT-L-SPM16 (ours). The token representations are colored in red, green and yellow to denote swing dancing, baking cookie, golf chipping classes, respectively. Tokens of the same class are distributed diffusely in the shallow layers ($4^{th}$, $8^{th}$, $12^{th}$) while are more concentrated in the deep layers ($16^{th}$, $20^{th}$, $24^{th}$). The margins produced by ViT with SPM among different classes are obviously larger and more explicit than those from the vanilla MAE-ViT-L in deep and shallow layers. Right: Accuracy-efficiency trade-off on K400 dataset.}
    \vspace{-0.5cm}
    \label{fig:_fig_tk_dist_cmp1}
\end{figure*}

\noindent \textbf{Visualization of Attention.} We provide examples for attention comparison between MAE-ViT-L (baseline) and MAE-ViT-L-SPM16 (ours) in Fig.\ref{fig:iccv_attn_vis}. In each video example, we show the raw video frames, the averaged attention score for each patch/token in the $16^{th}$ layer of the baseline, and the averaged semantic score for each patch/token in the $16^{th}$ layer of our approach (during SPM) in the first, second, and third rows, respectively.
The redder the color, the higher attention is placed on it. From these examples, it can be clearly observed that significant attention is placed on the background tokens and only small parts of the core objects/actors are covered in red color by the baseline. In contrast, our approach can focus more on the core objects and properly mute some less important background patches, indicating that, after our proposed SPM, the ratio of foreground tokens is successfully improved and the unnecessary background information is properly deemphasized.

\begin{figure}[t]
  \centering
   \includegraphics[width=1.0\linewidth]{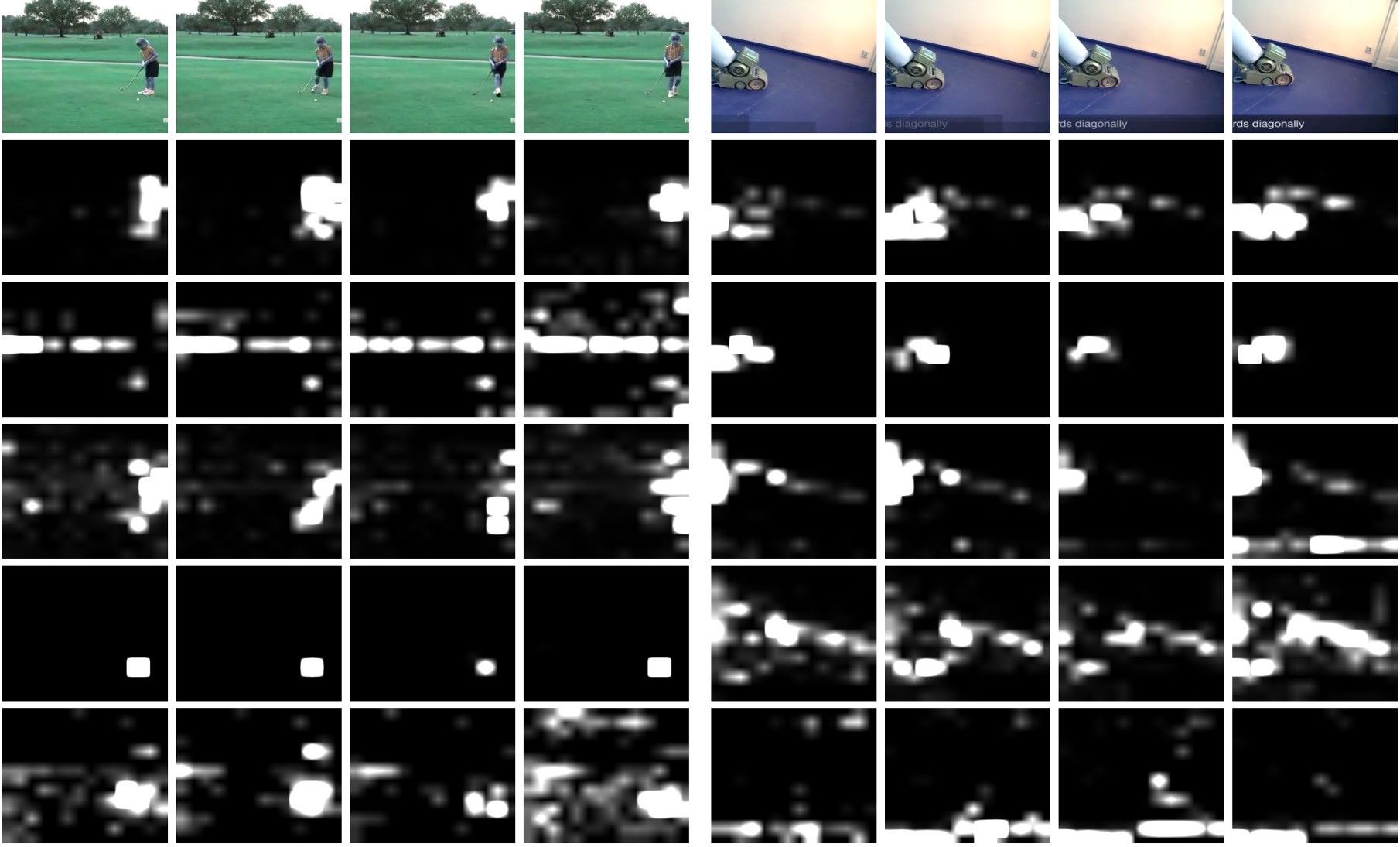}
   \vspace{-0.7cm}
   \caption{Semantic pooling visualization of MAE-ViT-L-SPM/16 for `golf chipping' and 'sanding floor' classes. The rows below the raw videos show the token content under one semantic prototype. We demonstrate 5 semantic pools from the total of 32 pools (remaining 27 pools are shown in the supplementary material.)}
   \label{fig:_fig_sempool_visx2}
   \vspace{-0.7cm}
\end{figure}

\noindent \textbf{Visualization of Semantic Pooling.} Fig.~\ref{fig:_fig_sempool_visx2} provides visualization of semantic pooling in MAE-ViT-L-SPM16. In each example, the first row shows the raw video and the following rows show the token content in five different semantic pools (we provide all the semantic visualizations in the supplementary material). The tokens located in the white or gray areas are aggregated together with the assigned weights in each cluster (the whiter it is the larger attention is given to the corresponding token), while the tokens located in the black area are removed from each cluster. 
For `golf chipping', the first and third pools collect the human information, the second pool collects the background tree information, the fourth pool collects the golf tokens, and the last pool captures the whole scene for the video and the golf-related parts are emphasized.
For `Sanding floor’ class, the first three pools mainly collect the tokens representing the sander, which is the core active object, while the last two pools collect the background features.
It is obvious that, before SPM, the core object occupies only a small portion of the whole video content, while after SPM, there are more semantic tokens describing the core objects and few representing the less important background.
Therefore, these examples clearly illustrate that our proposed SPM properly mitigates the issues stated in Sec.~\ref{sec:Introduction}, that is it can improve the token proportion of the core object and guide the model to focus more on the foreground part, which is helpful for understanding the video content. It also demonstrates that in our SPM, both foreground and background information are retained, meaning that it can adjust the token ratio without losing information.

\noindent \textbf{Visualization of Token Distribution.} We demonstrate the latent token representation distribution from the $4^{th}$, $8^{th}$, $12^{th}$, $16^{th}$, $20^{th}$, and $24^{th}$ layers of the baseline (single-scale MAE-ViT-L) and our MAE-ViT-L-SPM16 in Fig.~\ref{fig:_fig_tk_dist_cmp1}. The red, green and orange colors represent `swing dancing', `baking cookie' and `golf chipping' classes, respectively. While the baseline processes a total of 1568 tokens throughout the entire network, we apply the SPM at the 16th layer in our model to reduce the number of tokens to 128 for all subsequent layers.
With less color mixing area and larger margins between clusters with different classes in the shallow and deep layers, our model shows better representation ability than the baseline.
In addition, such superiority is achieved with much less number of tokens, which implies that our model can effectively represent the video content in a more memory-efficient way.

\vspace{-0.2cm}
\subsection{Ablation Studies} \label{ssec:ablations}
\vspace{-0.2cm}
In this section, we present part of our ablation studies on different configurations for integrating proposed SPM with MAE-ViT and MViTv2 in video recognition. 
For video recognition, we conduct all experiments with input size of $16 \times 224 \times 224 \times 3$ and $32 \times 224 \times 224 \times 3$ for MAE-ViT/MViTv2-S and MViT-B respectively.

\noindent \textbf{Threshold.} We investigate the thresholds in SPM, when integrated with MAE-ViT and MViTv2 on Kinetics-400. As shown in Tab.~\ref{tab:abl_th}, 0.7 is the best value for MAE-ViT-L and 0.5 is the best value for MViTv2-S. 

\noindent \textbf{Window size.} We investigated four window sizes while maintaining a constant 128 semantic tokens for SPM, when integrated with MAE-ViT-L and MViT-v2, on Kinetics-400 dataset. The results are shown in Tab.~\ref{tab:abl_window_size}. For MAE-ViT-L, we incorporate SPM in a single-pool way, where the SPM is inserted at the $16^{th}$ layer. As can be seen, window sizes of $2 \times 14 \times 14$ and $4 \times 7 \times 7$ perform best for MAE-ViT-L and MViTv2-S, respectively. 

\noindent \textbf{Methods of building MAE-ViT-SPM in a hierarchical way.} We progressively reduce the total number of tokens from 1568 to 1024, 512, 128 when incorporating SPM with MAE-ViT in a hierarchical way. To compare with the incorporation approach described in Sec.~\ref{sec:SPM_Model_Arch} and further prove the effectiveness of our module, we conduct experiments to progressively reduce the number of tokens with other strategies: (i) applying average/max pooling to keep only the original tokens; (ii) using SPM to keep only the semantic tokens; (iii) the combination method described in Sec.~\ref{sec:SPM_Model_Arch}. As shown in Tab.~\ref{tab:abl_SPM_VIT_hierarchical}, only the combination approach (Token$_{ori+sem}$) can surpass (85\%) the MAE-ViT-L baseline (84.8\%) on K400 dataset, illustrating the effectiveness of our strategy.
%
%

\begin{table}[t]
\centering
    \resizebox{0.6\linewidth}{!}{
        \begin{tabular}{l|ccc}
            \toprule
            model-dataset & 0.3 & 0.5 & 0.7 \\
            \midrule
            MAE-ViT-L & 84.6 & 84.9 & \textbf{85.1} \\
            MViTv2-S & 79.3 & \textbf{80.9} & 80.8 \\
            \bottomrule
        \end{tabular}
    }
\caption{Ablation studies for different threshold choices in SPM. MAE-ViT-L indicates the MAE-ViT-L-SPM16 with window size of $2 \times 14 \times 14$, and MViTv2-S indicates MViTv2-S-SPM with window size of $4 \times 7 \times 7$.}
\label{tab:abl_th}
\end{table}

\begin{table}[t]
\centering
    \resizebox{0.55\linewidth}{!}{
        \begin{tabular}{l|cc}
            \toprule
            model & top1 & top5 \\
            \midrule
            AvgPool & 83.7 & 96.0 \\
            MaxPool & 84.4 & 96.2 \\
            Token$_{sem}$ & 84.6 & 96.4 \\
            Token$_{ori+sem}$ & \textbf{85.0} & \textbf{96.5} \\
            \bottomrule
        \end{tabular}
    }
\caption{Ablation studies for different approaches to progressively reduce the number of tokens from 1568 to 1024, 512, 128 in layers 8, 14, and 18 respectively. Token$_{sem}$ and Token$_{ori+sem}$ indicate the method of applying SPM to keep only semantic tokens and the combination method described in Sec.~\ref{sec:SPM_Model_Arch}, respectively.}
\label{tab:abl_SPM_VIT_hierarchical}
\end{table}

\begin{table}[t]
\centering
    \resizebox{1.0\linewidth}{!}{
        \begin{tabular}{l|cccc}
            \toprule
            model & 4$\times$7$\times$7 & 1$\times$14$\times$14 & 2$\times$14$\times$14 & 8$\times$14$\times$14 \\
            \midrule
            MAE-ViT-L & 84.7 & 84.8 & \textbf{85.1} & 84.8 \\
            MViTv2-S & \textbf{80.9} & 80.6 & 80.0 & 80.5 \\
            \bottomrule
        \end{tabular}
    }
\caption{Ablation studies for different window size choices while maintaining a total number of 128 semantic tokens in SPM on K400 dataset. MAE-ViT-L indicates the MAE-ViT-L-SPM16 with window size of $2 \times 14 \times 14$ and threshold of 0.7, and MViTv2-S indicates MViTv2-S-SPM with window size of $4 \times 7 \times 7$ and threshold of 0.5.}
\label{tab:abl_window_size}
\end{table}


\noindent \textbf{Additional ablations.} To fully investigate the Semantic Pooling Module (SPM), we have conducted additional ablation studies on the following aspects with MAE-ViT-B on the K400 dataset:
(i) We study two methods for grouping tokens representing similar semantics. The first one is grouping neighbouring tokens based on the semantic distances. We sort the tokens within each window based on the semantic scores under each prototype. Then, we split them into $K$ groups based on the sorting order, and apply softmax on the semantic scores within each group. The module will finally generate $N_{r} = M \times K \times N_{win}$ semantic tokens after performing weighted sum between the tokens and normalized scores in each group. The second one is the elitism filtering as described in Sec.3.2.
(ii) We study the multi-head semantic pooling. When it is applied, we perform semantic pooling in parallel with multiple heads.
(iii) We compare the performances of applying output project layer and not using output project layer in SPM.
As can be seen from Tab.~\ref{tab:abl_base}, the performances of elitism approach are better and more stable than the neighboring approach. 

\begin{table}
    \centering
    \resizebox{1.0\linewidth}{!}{
    \begin{tabular}{l|cc|cccccccc}
        \toprule
model & top1 & top5 & \#head & M & K & $T_w \times H_w \times W_w$ & $N_{win}$ & L & O  \\
\midrule
neighbor & 80.38 & 94.39 & 12M & 16S & 8K & 8,14,14 & 1W & 8l & N  \\
neighbor & 80.16 & 94.42 & 12M & 8S & 8K & 4,14,14 & 2W & 8l & N  \\
neighbor & 79.73 & 94.22 & 12M & 8S & 4K & 2,14,14 & 4W & 8l & N  \\
neighbor & 79.66 & 94.27 & 12M & 4S & 4K & 1,14,14 & 8W & 8l & N  \\
neighbor & 80.11 & 94.22 & 12M & 32S & 4K & 8,14,14 & 1W & 8l & N  \\
neighbor & 79.66 & 93.87 & 12M & 8S & 2K & 1,14,14 & 8W & 8l & N  \\
neighbor & 80.08 & 94.31 & 16M & 16S & 8K & 8,14,14 & 1W & 8l & N  \\
neighbor & 80.54 & 94.32 & 8M & 16S & 8K & 8,14,14 & 1W & 8l & N  \\
neighbor & 80.00 & 94.27 & 1M & 16S & 8K & 8,14,14 & 1W & 8l & N  \\
neighbor & 77.10 & 92.8 & 12M & 16S & 8K & 8,14,14 & 1W & 6l & N  \\
neighbor & 72.25 & 89.9 & 12M & 16S & 8K & 8,14,14 & 1W & 4l & N  \\
neighbor & 79.50 & 94.00 & 12M & 16S & 8K & 8,14,14 & 1W & 8l & N  \\
\midrule
        \toprule
model & top1 & top5 & \#head & M & th & $T_w \times H_w \times W_w$ & $N_{win}$ & L & O  \\
\midrule
elitism & 80.80 & 94.82 & 1M & 128S & 0.7 & 8,14,14 & 1W & 8l & N  \\
elitism & 80.77 & 94.52 & 1M & 64S & 0.7 & 4,14,14 & 2W & 8l & N  \\
elitism & 80.85 & 94.84 & 1M & 32S & 0.7 & 2,14,14 & 4W & 8l & N  \\
elitism & 80.27 & 94.48 & 1M & 16S & 0.7 & 1,14,14 & 8W & 8l & N  \\
elitism & 80.65 & 94.62 & 1M & 128S & 0.5 & 8,14,14 & 1W & 8l & N  \\
elitism & 80.70 & 94.59 & 1M & 128S & 0.6 & 8,14,14 & 1W & 8l & N  \\
elitism & 80.57 & 94.74 & 1M & 128S & 0.9 & 8,14,14 & 1W & 8l & N  \\
elitism & 79.94 & 94.48 & 1M & 128S & 0.7 & 8,14,14 & 1W & 6l & N  \\
elitism & 78.10 & 93.37 & 1M & 128S & 0.7 & 8,14,14 & 1W & 4l & N  \\
elitism & 79.36 & 93.73 & 12M & 128S & 0.7 & 8,14,14 & 1W & 8l & N  \\
elitism & 80.54 & 94.57 & 1M & 128S & 0.7 & 8,14,14 & 1W & 8l & Y  \\
elitism & 80.73 & 94.74 & 1M & 128S & 0.7 & 8,14,14 & 1W & 8l & N  \\
        \bottomrule
    \end{tabular}
    }
    \caption{Ablation studies for MAE-VIT-B-SPM on K400. O indicates the output project layer. All experiments are based on a total of 128 supertokens.}
    \label{tab:abl_base}
    \vspace{-0.4cm}
\end{table}

\section{Conclusion}
\vspace{-0.2cm}
In this paper, we have presented a Supertoken Video Transformer, SVT, which employs our proposed semantic pooling module (SPM). SPM can be used with both single-scale and multi-scale transformers to reduce memory and computation requirements as well as improve the performance for video understanding. Thanks to adjusting the token proportion in the video pool and aggregating tokens based on semantics, our proposed module can reduce the video input redundancy without losing information, which has also been demonstrated via extensive experiments. With less computation and larger throughput, our model surpasses or provides comparable performance to both baselines and previous efficiency-oriented techniques on several datasets and vision tasks. In contrast to other computation/memory saving techniques, which can only be used by ViT and sacrifice performance, our module is applicable for both single-scale and multi-scale transformers without sacrificing  performance, which also proves the superiority of our SPM. 
Since, in this work, our main target was video understanding, we will further investigate the SPM on images in our future work.

\section*{Acknowledgments}
This work was done during an internship at Meta AI. Thanks to Chao-Yuan Wu and Christoph Feichtenhofer for helpful feedback.




\bibliographystyle{IEEEtran}
\bibliography{egbib}

\begin{thebibliography}{10}
\providecommand{\url}[1]{#1}
\csname url@samestyle\endcsname
\providecommand{\newblock}{\relax}
\providecommand{\bibinfo}[2]{#2}
\providecommand{\BIBentrySTDinterwordspacing}{\spaceskip=0pt\relax}
\providecommand{\BIBentryALTinterwordstretchfactor}{4}
\providecommand{\BIBentryALTinterwordspacing}{\spaceskip=\fontdimen2\font plus
\BIBentryALTinterwordstretchfactor\fontdimen3\font minus
  \fontdimen4\font\relax}
\providecommand{\BIBforeignlanguage}[2]{{%
\expandafter\ifx\csname l@#1\endcsname\relax
\typeout{** WARNING: IEEEtran.bst: No hyphenation pattern has been}%
\typeout{** loaded for the language `#1'. Using the pattern for}%
\typeout{** the default language instead.}%
\else
\language=\csname l@#1\endcsname
\fi
#2}}
\providecommand{\BIBdecl}{\relax}
\BIBdecl

\bibitem{vaswani2017attention}
A.~Vaswani, N.~Shazeer, N.~Parmar, J.~Uszkoreit, L.~Jones, A.~N. Gomez,
  {\L}.~Kaiser, and I.~Polosukhin, ``Attention is all you need,''
  \emph{Advances in neural information processing systems}, vol.~30, 2017.

\bibitem{devlin2018bert}
J.~Devlin, M.-W. Chang, K.~Lee, and K.~Toutanova, ``Bert: Pre-training of deep
  bidirectional transformers for language understanding,'' \emph{arXiv preprint
  arXiv:1810.04805}, 2018.

\bibitem{arnab2021vivit}
A.~Arnab, M.~Dehghani, G.~Heigold, C.~Sun, M.~Lu{\v{c}}i{\'c}, and C.~Schmid,
  ``Vivit: A video vision transformer,'' in \emph{Proceedings of the IEEE/CVF
  International Conference on Computer Vision}, 2021, pp. 6836--6846.

\bibitem{dosovitskiy2020image}
A.~Dosovitskiy, L.~Beyer, A.~Kolesnikov, D.~Weissenborn, X.~Zhai,
  T.~Unterthiner, M.~Dehghani, M.~Minderer, G.~Heigold, S.~Gelly \emph{et~al.},
  ``An image is worth 16x16 words: Transformers for image recognition at
  scale,'' \emph{arXiv preprint arXiv:2010.11929}, 2020.

\bibitem{wu2022tinyvit}
K.~Wu, J.~Zhang, H.~Peng, M.~Liu, B.~Xiao, J.~Fu, and L.~Yuan, ``Tinyvit: Fast
  pretraining distillation for small vision transformers,'' in \emph{Computer
  Vision--ECCV 2022: 17th European Conference, Tel Aviv, Israel, October
  23--27, 2022, Proceedings, Part XXI}.\hskip 1em plus 0.5em minus 0.4em\relax
  Springer, 2022, pp. 68--85.

\bibitem{cho2022cross}
J.~Cho, K.~Youwang, and T.-H. Oh, ``Cross-attention of disentangled modalities
  for 3d human mesh recovery with transformers,'' in \emph{Computer
  Vision--ECCV 2022: 17th European Conference, Tel Aviv, Israel, October
  23--27, 2022, Proceedings, Part I}.\hskip 1em plus 0.5em minus 0.4em\relax
  Springer, 2022, pp. 342--359.

\bibitem{gao2022aiatrack}
S.~Gao, C.~Zhou, C.~Ma, X.~Wang, and J.~Yuan, ``Aiatrack: Attention in
  attention for transformer visual tracking,'' in \emph{Computer Vision--ECCV
  2022: 17th European Conference, Tel Aviv, Israel, October 23--27, 2022,
  Proceedings, Part XXII}.\hskip 1em plus 0.5em minus 0.4em\relax Springer,
  2022, pp. 146--164.

\bibitem{ye2022joint}
B.~Ye, H.~Chang, B.~Ma, S.~Shan, and X.~Chen, ``Joint feature learning and
  relation modeling for tracking: A one-stream framework,'' in \emph{Computer
  Vision--ECCV 2022: 17th European Conference, Tel Aviv, Israel, October
  23--27, 2022, Proceedings, Part XXII}.\hskip 1em plus 0.5em minus 0.4em\relax
  Springer, 2022, pp. 341--357.

\bibitem{yan2022towards}
B.~Yan, Y.~Jiang, P.~Sun, D.~Wang, Z.~Yuan, P.~Luo, and H.~Lu, ``Towards grand
  unification of object tracking,'' in \emph{Computer Vision--ECCV 2022: 17th
  European Conference, Tel Aviv, Israel, October 23--27, 2022, Proceedings,
  Part XXI}.\hskip 1em plus 0.5em minus 0.4em\relax Springer, 2022, pp.
  733--751.

\bibitem{zhao2022tracking}
Z.~Zhao, Z.~Wu, Y.~Zhuang, B.~Li, and J.~Jia, ``Tracking objects as pixel-wise
  distributions,'' in \emph{Computer Vision--ECCV 2022: 17th European
  Conference, Tel Aviv, Israel, October 23--27, 2022, Proceedings, Part
  XXII}.\hskip 1em plus 0.5em minus 0.4em\relax Springer, 2022, pp. 76--94.

\bibitem{dong2022cswin}
X.~Dong, J.~Bao, D.~Chen, W.~Zhang, N.~Yu, L.~Yuan, D.~Chen, and B.~Guo,
  ``Cswin transformer: A general vision transformer backbone with cross-shaped
  windows,'' in \emph{Proceedings of the IEEE/CVF Conference on Computer Vision
  and Pattern Recognition}, 2022, pp. 12\,124--12\,134.

\bibitem{pu2022edter}
M.~Pu, Y.~Huang, Y.~Liu, Q.~Guan, and H.~Ling, ``Edter: Edge detection with
  transformer,'' in \emph{Proceedings of the IEEE/CVF Conference on Computer
  Vision and Pattern Recognition}, 2022, pp. 1402--1412.

\bibitem{wang2022bridged}
Y.~Wang, T.~Ye, L.~Cao, W.~Huang, F.~Sun, F.~He, and D.~Tao, ``Bridged
  transformer for vision and point cloud 3d object detection,'' in
  \emph{Proceedings of the IEEE/CVF Conference on Computer Vision and Pattern
  Recognition}, 2022, pp. 12\,114--12\,123.

\bibitem{zhou2021deepvit}
D.~Zhou, B.~Kang, X.~Jin, L.~Yang, X.~Lian, Z.~Jiang, Q.~Hou, and J.~Feng,
  ``Deepvit: Towards deeper vision transformer,'' \emph{arXiv preprint
  arXiv:2103.11886}, 2021.

\bibitem{ni2022expanding}
B.~Ni, H.~Peng, M.~Chen, S.~Zhang, G.~Meng, J.~Fu, S.~Xiang, and H.~Ling,
  ``Expanding language-image pretrained models for general video recognition,''
  in \emph{Computer Vision--ECCV 2022: 17th European Conference, Tel Aviv,
  Israel, October 23--27, 2022, Proceedings, Part IV}.\hskip 1em plus 0.5em
  minus 0.4em\relax Springer, 2022, pp. 1--18.

\bibitem{li2022mvitv2}
Y.~Li, C.-Y. Wu, H.~Fan, K.~Mangalam, B.~Xiong, J.~Malik, and C.~Feichtenhofer,
  ``Mvitv2: Improved multiscale vision transformers for classification and
  detection,'' in \emph{Proceedings of the IEEE/CVF Conference on Computer
  Vision and Pattern Recognition}, 2022, pp. 4804--4814.

\bibitem{liu2021video}
Z.~Liu, J.~Ning, Y.~Cao, Y.~Wei, Z.~Zhang, S.~Lin, and H.~Hu, ``Video swin
  transformer,'' \emph{arXiv preprint arXiv:2106.13230}, 2021.

\bibitem{neimark2021video}
D.~Neimark, O.~Bar, M.~Zohar, and D.~Asselmann, ``Video transformer network,''
  in \emph{Proceedings of the IEEE/CVF International Conference on Computer
  Vision}, 2021, pp. 3163--3172.

\bibitem{girdhar2021anticipative}
R.~Girdhar and K.~Grauman, ``Anticipative video transformer,'' in
  \emph{Proceedings of the IEEE/CVF international conference on computer
  vision}, 2021, pp. 13\,505--13\,515.

\bibitem{zhang2021vidtr}
Y.~Zhang, X.~Li, C.~Liu, B.~Shuai, Y.~Zhu, B.~Brattoli, H.~Chen, I.~Marsic, and
  J.~Tighe, ``Vidtr: Video transformer without convolutions,'' in
  \emph{Proceedings of the IEEE/CVF international conference on computer
  vision}, 2021, pp. 13\,577--13\,587.

\bibitem{bulat2021space}
A.~Bulat, J.~M. Perez~Rua, S.~Sudhakaran, B.~Martinez, and G.~Tzimiropoulos,
  ``Space-time mixing attention for video transformer,'' \emph{Advances in
  Neural Information Processing Systems}, vol.~34, pp. 19\,594--19\,607, 2021.

\bibitem{liang2022vrt}
J.~Liang, J.~Cao, Y.~Fan, K.~Zhang, R.~Ranjan, Y.~Li, R.~Timofte, and
  L.~Van~Gool, ``Vrt: A video restoration transformer,'' \emph{arXiv preprint
  arXiv:2201.12288}, 2022.

\bibitem{ranasinghe2022self}
K.~Ranasinghe, M.~Naseer, S.~Khan, F.~S. Khan, and M.~S. Ryoo,
  ``Self-supervised video transformer,'' in \emph{Proceedings of the IEEE/CVF
  Conference on Computer Vision and Pattern Recognition}, 2022, pp. 2874--2884.

\bibitem{wang2022bevt}
R.~Wang, D.~Chen, Z.~Wu, Y.~Chen, X.~Dai, M.~Liu, Y.-G. Jiang, L.~Zhou, and
  L.~Yuan, ``Bevt: Bert pretraining of video transformers,'' in
  \emph{Proceedings of the IEEE/CVF Conference on Computer Vision and Pattern
  Recognition}, 2022, pp. 14\,733--14\,743.

\bibitem{zhang2021token}
H.~Zhang, Y.~Hao, and C.-W. Ngo, ``Token shift transformer for video
  classification,'' in \emph{Proceedings of the 29th ACM International
  Conference on Multimedia}, 2021, pp. 917--925.

\bibitem{girdhar2019video}
R.~Girdhar, J.~Carreira, C.~Doersch, and A.~Zisserman, ``Video action
  transformer network,'' in \emph{Proceedings of the IEEE/CVF conference on
  computer vision and pattern recognition}, 2019, pp. 244--253.

\bibitem{wang2022deformable}
J.~Wang and L.~Torresani, ``Deformable video transformer,'' in
  \emph{Proceedings of the IEEE/CVF Conference on Computer Vision and Pattern
  Recognition}, 2022, pp. 14\,053--14\,062.

\bibitem{kim2022tubeformer}
D.~Kim, J.~Xie, H.~Wang, S.~Qiao, Q.~Yu, H.-S. Kim, H.~Adam, I.~S. Kweon, and
  L.-C. Chen, ``Tubeformer-deeplab: Video mask transformer,'' in
  \emph{Proceedings of the IEEE/CVF Conference on Computer Vision and Pattern
  Recognition}, 2022, pp. 13\,914--13\,924.

\bibitem{shi2022video}
Z.~Shi, X.~Xu, X.~Liu, J.~Chen, and M.-H. Yang, ``Video frame interpolation
  transformer,'' in \emph{Proceedings of the IEEE/CVF Conference on Computer
  Vision and Pattern Recognition}, 2022, pp. 17\,482--17\,491.

\bibitem{herzig2022object}
R.~Herzig, E.~Ben-Avraham, K.~Mangalam, A.~Bar, G.~Chechik, A.~Rohrbach,
  T.~Darrell, and A.~Globerson, ``Object-region video transformers,'' in
  \emph{Proceedings of the IEEE/CVF Conference on Computer Vision and Pattern
  Recognition}, 2022, pp. 3148--3159.

\bibitem{liu2022learning}
C.~Liu, H.~Yang, J.~Fu, and X.~Qian, ``Learning trajectory-aware transformer
  for video super-resolution,'' in \emph{Proceedings of the IEEE/CVF Conference
  on Computer Vision and Pattern Recognition}, 2022, pp. 5687--5696.

\bibitem{fan2021multiscale}
H.~Fan, B.~Xiong, K.~Mangalam, Y.~Li, Z.~Yan, J.~Malik, and C.~Feichtenhofer,
  ``Multiscale vision transformers,'' in \emph{Proceedings of the IEEE/CVF
  International Conference on Computer Vision}, 2021, pp. 6824--6835.

\bibitem{liu2021swin}
Z.~Liu, Y.~Lin, Y.~Cao, H.~Hu, Y.~Wei, Z.~Zhang, S.~Lin, and B.~Guo, ``Swin
  transformer: Hierarchical vision transformer using shifted windows,'' in
  \emph{Proceedings of the IEEE/CVF International Conference on Computer
  Vision}, 2021, pp. 10\,012--10\,022.

\bibitem{feichtenhofer2022masked}
C.~Feichtenhofer, H.~Fan, Y.~Li, and K.~He, ``Masked autoencoders as
  spatiotemporal learners,'' \emph{arXiv preprint arXiv:2205.09113}, 2022.

\bibitem{xie2021so}
J.~Xie, R.~Zeng, Q.~Wang, Z.~Zhou, and P.~Li, ``So-vit: Mind visual tokens for
  vision transformer,'' \emph{arXiv preprint arXiv:2104.10935}, 2021.

\bibitem{yuan2021tokens}
L.~Yuan, Y.~Chen, T.~Wang, W.~Yu, Y.~Shi, Z.-H. Jiang, F.~E. Tay, J.~Feng, and
  S.~Yan, ``Tokens-to-token vit: Training vision transformers from scratch on
  imagenet,'' in \emph{Proceedings of the IEEE/CVF International Conference on
  Computer Vision}, 2021, pp. 558--567.

\bibitem{Touvron2021deit}
H.~Touvron, M.~Cord, M.~Douze, F.~Massa, A.~Sablayrolles, and H.~Jegou,
  ``Training data-efficient image transformers amp; distillation through
  attention,'' in \emph{ICCV}, 2021.

\bibitem{Neimark2021VTN}
M.~Z. Daniel~Neimark, Omri~Bar and D.~Asselmann, ``Video transformer network,''
  \emph{arXiv preprint arXiv:2102.00719}, 2021.

\bibitem{bertasius2021space}
G.~Bertasius, H.~Wang, and L.~Torresani, ``Is space-time attention all you need
  for video understanding?'' \emph{arXiv preprint arXiv:2102.05095}, 2021.

\bibitem{patrick2021keeping}
M.~Patrick, D.~Campbell, Y.~M. Asano, I.~M.~F. Metze, C.~Feichtenhofer,
  A.~Vedaldi, J.~Henriques \emph{et~al.}, ``Keeping your eye on the ball:
  Trajectory attention in video transformers,'' \emph{arXiv preprint
  arXiv:2106.05392}, 2021.

\bibitem{herzig2021object}
R.~Herzig, E.~Ben-Avraham, K.~Mangalam, A.~Bar, G.~Chechik, A.~Rohrbach,
  T.~Darrell, and A.~Globerson, ``Object-region video transformers,''
  \emph{arXiv preprint arXiv:2110.06915}, 2021.

\bibitem{kay2017kinetics}
W.~Kay, J.~Carreira, K.~Simonyan, B.~Zhang, C.~Hillier, S.~Vijayanarasimhan,
  F.~Viola, T.~Green, T.~Back, P.~Natsev \emph{et~al.}, ``The kinetics human
  action video dataset,'' \emph{arXiv preprint arXiv:1705.06950}, 2017.

\bibitem{goyal2017something}
R.~Goyal, S.~Ebrahimi~Kahou, V.~Michalski, J.~Materzynska, S.~Westphal, H.~Kim,
  V.~Haenel, I.~Fruend, P.~Yianilos, M.~Mueller-Freitag \emph{et~al.}, ``The"
  something something" video database for learning and evaluating visual common
  sense,'' in \emph{Proceedings of the IEEE international conference on
  computer vision}, 2017, pp. 5842--5850.

\bibitem{meng2022adavit}
L.~Meng, H.~Li, B.-C. Chen, S.~Lan, Z.~Wu, Y.-G. Jiang, and S.-N. Lim,
  ``Adavit: Adaptive vision transformers for efficient image recognition,'' in
  \emph{Proceedings of the IEEE/CVF Conference on Computer Vision and Pattern
  Recognition}, 2022, pp. 12\,309--12\,318.

\bibitem{yin2022vit}
H.~Yin, A.~Vahdat, J.~M. Alvarez, A.~Mallya, J.~Kautz, and P.~Molchanov,
  ``A-vit: Adaptive tokens for efficient vision transformer,'' in
  \emph{Proceedings of the IEEE/CVF Conference on Computer Vision and Pattern
  Recognition}, 2022, pp. 10\,809--10\,818.

\bibitem{rao2021dynamicvit}
Y.~Rao, W.~Zhao, B.~Liu, J.~Lu, J.~Zhou, and C.-J. Hsieh, ``Dynamicvit:
  Efficient vision transformers with dynamic token sparsification,''
  \emph{Advances in neural information processing systems}, vol.~34, pp.
  13\,937--13\,949, 2021.

\bibitem{kong2021spvit}
Z.~Kong, P.~Dong, X.~Ma, X.~Meng, W.~Niu, M.~Sun, B.~Ren, M.~Qin, H.~Tang, and
  Y.~Wang, ``Spvit: Enabling faster vision transformers via soft token
  pruning,'' \emph{arXiv preprint arXiv:2112.13890}, 2021.

\bibitem{ryoo2021tokenlearner}
M.~S. Ryoo, A.~Piergiovanni, A.~Arnab, M.~Dehghani, and A.~Angelova,
  ``Tokenlearner: What can 8 learned tokens do for images and videos?''
  \emph{arXiv preprint arXiv:2106.11297}, 2021.

\bibitem{bolya2022token}
D.~Bolya, C.-Y. Fu, X.~Dai, P.~Zhang, C.~Feichtenhofer, and J.~Hoffman, ``Token
  merging: Your vit but faster,'' \emph{arXiv preprint arXiv:2210.09461}, 2022.

\end{thebibliography}

\newpage

 


\vfill

\end{document}